\documentclass[11pt,a4paper,english]{article}
\usepackage[hyperref]{naaclhlt2018}
\usepackage{times}
\usepackage{latexsym}
\usepackage{babel}
\usepackage[pdftex]{graphicx}
\usepackage{footnote}
\usepackage{tablefootnote}
\usepackage{url}
\usepackage{enumitem}
\makesavenoteenv{table}

\usepackage{amssymb}% http://ctan.org/pkg/amssymb
\usepackage{pifont}% http://ctan.org/pkg/pifont
\usepackage[scaled=.7]{beramono}
\newcommand{\cmark}{\ding{51}}%
\newcommand{\xmark}{\ding{55}}%

\aclfinalcopy
\def\aclpaperid{574839}

\begin{document}

\title{Text Segmentation as a Supervised Learning Task}
\author{Omri Koshorek\thanks{ \hspace{0.15cm}Both authors contributed equally to this paper and the order of authorship was determined randomly.}\quad Adir Cohen\footnotemark[1] \quad Noam Mor \quad Michael Rotman \quad Jonathan Berant\\
School of Computer Science \\
Tel-Aviv University, Israel \\
\texttt{\{omri.koshorek,adir.cohen,noam.mor,michael.rotman,joberant\}@cs.tau.ac.il} }
\maketitle

\begin{abstract}
Text segmentation, the task of dividing a document into contiguous segments based on its semantic structure, is a longstanding challenge in language understanding. Previous work on text segmentation focused on unsupervised methods such as clustering or graph search, due to the paucity in labeled data. In this work, we formulate text segmentation as a supervised learning problem, and present a  large new dataset for text segmentation that is automatically extracted and labeled from Wikipedia. Moreover, we develop a segmentation model based on this dataset and show that it generalizes well to unseen natural text.
\end{abstract}

\section{Introduction}
\label{sec:intro}
Text segmentation is the task of dividing text into segments, such that each segment is topically coherent, and cutoff points indicate a change of topic \cite{hearst1994multi, utiyama2001statistical,brants2002topic}. 
This provides basic structure to a  document in a way that can later be used by downstream applications such as summarization and information extraction.

% \jb{Need to say something more explicit about the state of existing datasets for document segmentation}
Existing datasets for text segmentation are small in size \cite{choi2000advances,glavavs2016unsupervised}, and are used mostly for evaluating the performance of  segmentation algorithms. Moreover, some datasets \cite{choi2000advances} were synthesized automatically and thus do not represent the natural distribution of text in documents. 
%For example, the most popular benchmark is completely synthetic. This being the case, we strongly feel that a new benchmark for general text segmentation is required.
Because no large labeled dataset  exists, prior work on text segmentation  tried to either come up with heuristics for identifying whether two sentences discuss the same topic  \cite{choi2000advances,glavavs2016unsupervised}, or to model topics explicitly with methods such as LDA \cite{blei03lda} that assign a topic to each paragraph or sentence \cite{chen2009global}.

Recent developments in Natural Language Processing have demonstrated that casting problems as supervised learning tasks over large amounts of labeled data is highly effective compared to heuristic-based systems or unsupervised algorithms \cite{mikolov2013linguistic,pennington2014glove}. Therefore, in this work we (a) formulate text segmentation as a supervised learning problem, where a label for every sentence in the document denotes whether it ends a segment,
(b) describe a new dataset, \textsc{Wiki-727k}, intended for training text segmentation models.

\textsc{Wiki-727k} comprises more than 727,000 documents from English Wikipedia, where the table of contents of each document is used to automatically segment the document. Since this dataset is large, natural, and covers a variety of topics, we expect it to generalize well to other natural texts. Moreover, \textsc{Wiki-727k} provides a better benchmark for evaluating text segmentation models compared to existing datasets. We make \textsc{Wiki-727k} and our code publicly available at \url{https://github.com/koomri/text-segmentation}.

To demonstrate the efficacy of this dataset, we develop a hierarchical neural model in which a lower-level bidirectional LSTM creates sentence representations from word tokens, and then a higher-level LSTM consumes the sentence representations and labels each sentence.
We show that our model outperforms prior methods, demonstrating the importance of  our dataset for future progress in text segmentation. 

%for text segmentation, and show that it outperforms prior methods on ourOur new dataset provides both a training set and a benchmark that represent real-world text in a much better way than previous datasets. We believe this to be a necessary step in text segmentation research.

\section{Related Work}
\label{sec:rel}

% \jb{In general this section can be shorter I think.}

\subsection{Existing Text Segmentation Datasets}
The most common dataset for evaluating performance on text segmentation was created by \newcite{choi2000advances}. It is a synthetic dataset containing 920 documents, where each document is a concatenation of 10 random passages from the Brown corpus. 
\newcite{glavavs2016unsupervised} created a dataset of their own, which consists of 5 manually-segmented political manifestos from the Manifesto project.\footnote{\url{https://manifestoproject.wzb.eu}}
\cite{chen2009global} also used English Wikipedia documents to evaluate text segmentation. They defined two datasets, one with 100 documents about major cities and one with 118 documents about chemical elements. Table \ref{tab:statistics} provides additional statistics on each dataset.

Thus, all existing datasets for text segmentation are small and cannot benefit from the advantages of training supervised models over labeled data.

\subsection{Previous Methods}
Bayesian text segmentation methods \cite{chen2009global,riedl2012topictiling} employ a generative probabilistic model for text. In these models, a document is represented as a set of topics, which are sampled from a topic distribution, and each topic imposes a distribution over the vocabulary. \newcite{riedl2012topictiling} perform best among this family of methods, where they define a coherence score between pairs of sentences, and compute a segmentation by finding drops in coherence scores between pairs of adjacent sentences.

Another noteworthy approach for text segmentation is \textsc{GraphSeg} \cite{glavavs2016unsupervised}, an unsupervised graph method, which performs competitively on synthetic datasets and outperforms Bayesian approaches on the Manifesto dataset. \textsc{GraphSeg} works by building a graph where nodes are sentences, and an edge between two sentences signifies that the sentences are semantically similar. The segmentation is then determined by finding maximal cliques of adjacent sentences, and heuristically completing the segmentation. 

\section{The \textsc{Wiki-727k} Dataset}
For this work we have created a new dataset, which we name \textsc{Wiki-727k}. It is a collection of 727,746 English Wikipedia documents, and their hierarchical segmentation, as it appears in their table of contents. We randomly partitioned the documents into a train (80\%), development (10\%), and test (10\%) set.

Different text segmentation use-cases  require different levels of granularity. For example, for segmenting text by overarching topic it makes sense to train a model that predicts only top-level segments, which are typically vary in topic -- for example, \emph{``History"}, \emph{``Geography"}, and \emph{``Demographics"}. For segmenting a radio broadcast into separate news stories, which requires finer granularity, it  makes sense to train a model to predict sub-segments. Our dataset provides the entire segmentation information, and an application may choose the appropriate level of granularity.

To generate the data, we performed the following preprocessing steps for each Wikipedia document:

\begin{itemize}[noitemsep]
\item Removed all photos, tables, Wikipedia template elements, and other non-text elements.
\item Removed single-sentence segments, documents with less than three segments, and documents where most segments were filtered.
\item Divided each segment into sentences using the \textsc{Punkt} tokenizer of the NLTK library \cite{bird2009nltk}. This is  necessary for the use of our dataset as a benchmark, as without a well-defined sentence segmentation, it is impossible to evaluate different models.
\end{itemize}

We view \textsc{Wiki-727k} as suitable for text segmentation because it is natural, open-domain, and has a well-defined segmentation. Moreover, neural network models often benefit from a wealth of training data, and our dataset can easily be further expanded at very little cost.

\section{Neural Model for Text Segmentation}
\label{sec:model}
%Both modules are based on 

We treat text segmentation as a supervised learning task, where the input $x$ is a document, represented as a sequence of $n$ sentences $s_1, \ldots, s_n$, and the label $y = (y_1, \dots, y_{n-1})$ is a segmentation of the document, represented by $n-1$ binary values, where $y_i$ denotes whether $s_i$ ends a segment.

We now describe our model for text segmentation. 
Our neural model is composed of a hierarchy of two sub-networks, both based on the LSTM architecture \cite{hochreiter1997long}. The lower-level sub-network is a two-layer bidirectional LSTM that generates sentence representations: for each sentence $s_i$, the network
consumes the words $w_1^{(i)}, \ldots, w_k^{(i)}$ of $s_i$ one by one, and 
the final sentence representation $e_i$ is computed by max-pooling over the LSTM outputs.
%takes the sentence as a sequence of words, and inputs their embeddings $w_1 \ldots w_k$ to a 2-layer bidirectional LSTM. 
%The sentence embedding $e_i$ is then computed as the result of max-pooling over the LSTM outputs.

\begin{figure}
\begin{center}
\includegraphics[width=50mm, height=60mm]{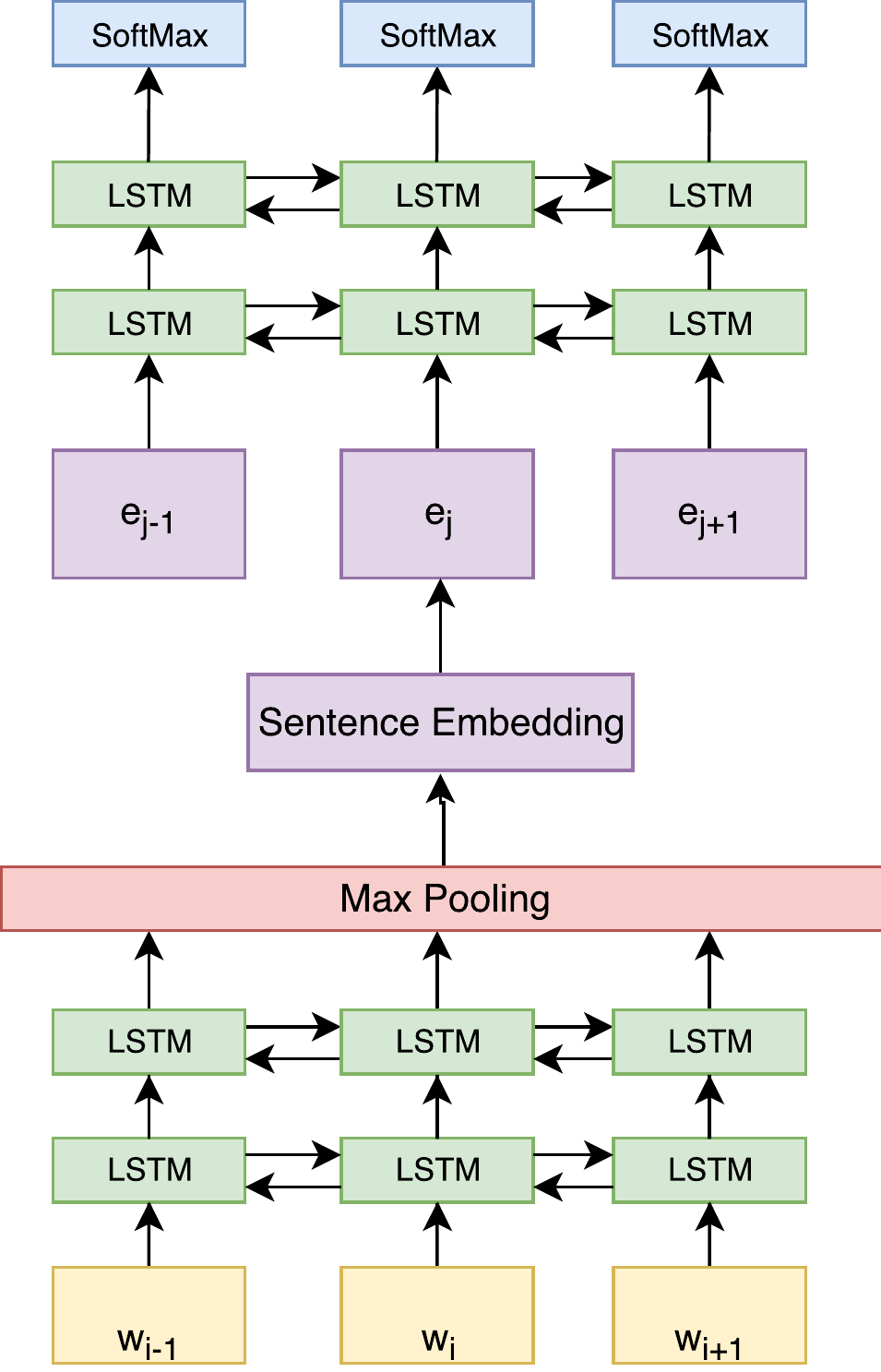}
\caption{Our model contains a sentence embedding sub-network, followed by a segmentation prediction sub-network which predicts a cut-off probability for each sentence.}
\label{model-architecture-figure}
\label{model}

\end{center}

\end{figure}

The higher-level sub-network is the segmentation prediction network. This sub-network takes a sequence of sentence embeddings $e_1, \ldots, e_n$ as input, and feeds them into a two-layer bidirectional LSTM. We then apply a fully-connected layer on each of the LSTM outputs to obtain a sequence of 
$n$ vectors in $\mathbb{R}^2$. 
We ignore the last vector (for $e_n$), and apply a \textit{softmax} function to obtain $n-1$ segmentation probabilities.  Figure~\ref{model-architecture-figure} illustrates the overall neural network architecture.

%$n$ vectors in $\mathbb{R}^2$. We ignore the last vector (for $e_n$), and apply a \textit{softmax} function to obtain $n-1$ segmentation probabilities $p_i$.  Figure~\ref{model-architecture-figure} illustrates the overall neural network's architecture.

% \subsection{Model variants}
% We have tested several variants of the sentence embedding module. These are the best performers:
% \begin{itemize}
% \item Last State model: Uses a bidirectional LSTM for both modules, and computes the sentence embeddings as the concatenation of the last hidden state vectors.

% \item Max sentence embedding: The sentence embedding is the result of a maximum operation on the last bi-LSTM output in the sentence embedding module.
% \end{itemize}

% It is worth noting that single-directional LSTM models performed very badly.

\subsection{Training}
Our model predicts for each sentence $s_i$, the probability $p_i$ that it ends a segment. For an $n$-sentence document, 
we minimize the sum of cross-entropy errors over each of the $n-1$ relevant sentences:
\[
J(\Theta) = \sum_{i=1}^{n-1}\left[-y_{i}\log p_{i}-\left(1-y_{i}\right)\log\left(1-p_{i}\right)\right].
\]
Training is done by stochastic gradient descent in an end-to-end manner. For word embeddings, we use the GoogleNews word2vec
%\cite{gword2vec} 
pre-trained model.

We train our system to only predict the top-level segmentation (other granularities are possible).
%This is purely a choice for training time, and not a limitation of the dataset. 
In addition, at training time, we removed from each document the first segment, since in Wikipedia it is often a summary that touches many different topics, and is therefore less useful for training a segmentation model. We also omitted lists and code snippets tokens.

\subsection{Inference}
At test time, the model takes a sequence of word embeddings divided
into sentences, and returns a vector $p$ of cutoff probabilities between
sentences. We use greedy decoding, i.e., we create a new segment
whenever $p_{i}$ is greater than a threshold $\tau$. We optimize the parameter $\tau$ on our validation set, and use the optimal value while testing.

\begin{table*}[t]
\centering

\label{datasets-details-table}
\caption{Statistics on various text segmentation datasets.}
{\footnotesize
\begin{tabular}{|l|c|c|c|c|c|}
\hline
 & \textsc{Wiki-727k} & \textsc{Choi} & \textsc{Manifesto} & \textsc{Cities}  & \textsc{Elements}\\ \hline
Documents & 727,746 & 920 & 5 & 100 & 118 \\ \hline
Segment Length\tablefootnote{Statistics on top-level segments.\label{tl-fn}} & 13.6 $\pm$ 20.3  & 7.4 $\pm$ 2.96 & 8.99 $\pm$ 10.8 & 5.15 $\pm$ 4.57 & 3.33 $\pm$ 3.05 \\  \hline
Segments per document\footnotemark[\getrefnumber{tl-fn}] & 3.48 $\pm$ 2.23 & 9.98 $\pm$ 0.12 & 127 $\pm$ 42.9 & 12.2 $\pm$ 2.79 & 6.82 $\pm$ 2.57 \\  \hline
Real-world & \cmark & \xmark & \cmark & \cmark & \cmark  \\ \hline
Large variety of topics & \cmark & \xmark & \xmark & \xmark & \xmark \\ \hline

\end{tabular}
}
\label{tab:statistics}
\end{table*}

\section{Experimental Details}
\label{sec:eval}
% \jb{Should write explicitly (a) what are the baselines you are comparing to. (b) what are the evaluation metrics you are considering}

We evaluate our method on the \textsc{Wiki-727} test set, Choi's synthetic dataset, and the two small Wikipedia datasets (\textsc{Cities}, \textsc{Elements}) introduced by \newcite{chen2009global}. We compare our model performance with those reported by \newcite{chen2009global} and \textsc{GraphSeg}. In addition, we evaluate the performance of a random baseline model, which starts a new segment after every sentence with probability $\frac{1}{k}$, where $k$ is the average segment size in the dataset.

Because our test set is large, it is difficult to evaluate some of the existing methods, which are computationally demanding.
Thus, we introduce \textsc{Wiki-50}, a set of 50 randomly sampled test documents from \textsc{Wiki-727K}. We use \textsc{Wiki-50} to evaluate systems that are too slow to evaluate on the entire test set. We also provide human segmentation performance results on \textsc{Wiki-50}.

We use the $P_k$ metric as defined in \newcite{beeferman1999statistical} to evaluate the performance of our model. $P_k$ is the probability that when passing a sliding window of size $k$ over \textit{sentences}, the sentences at the boundaries of the window will be incorrectly classified as belonging to the same segment (or vice versa). To match the setup of \newcite{chen2009global}, we also provide the $P_k$ metric for a sliding window over \textit{words} when evaluating on the datasets from their paper. Following \cite{glavavs2016unsupervised}, we set $k$ to half of the average segment size in the ground-truth segmentation. For evaluations we used the \textsc{SegEval} package \cite{Fournier2013a}.

In addition to segmentation accuracy, we also report runtime when running on a mid-range laptop CPU.

% \subsection{The $P_k$ Segmentation Quality Metric}

% \subsection{Comparisons}

\begin{table*}[t!]
\centering
{\footnotesize
\label{pk-table}
\caption{$P_k$ Results on the test set.}
\begin{tabular}{|l|c|c|c|cc|cc|}

\hline
 & \textsc{Wiki-727k} & \textsc{Wiki-50} & \textsc{Choi} & \multicolumn{2}{c|}{\textsc{Cities}} & \multicolumn{2}{c|}{\textsc{Elements}} \\ 
 $P_k$ variant&\textit{sentences}&\textit{sentences}&\textit{sentences}&\textit{sentences}&\textit{words}&\textit{sentences}&\textit{words}\\ \hline\hline
\cite{chen2009global} &   -  & -& -& -& 22.1& - & \textbf{20.1}\\ \hline
GraphSeg & - & 63.56 & \textbf{5.6-7.2}  & 39.95 &-& 49.12 &- \\ \hline
Our model & 22.13 & \textbf{18.24} & 26.26\tablefootnote{We optimized $\tau$ by cross validation on the \textsc{Choi}  dataset.}  & \textbf{19.68} & \textbf{18.14}& \textbf{41.63} & 33.82\\ \hline \hline
Random baseline & 53.09 & 52.65 & 49.43 & 47.14 & 44.14 & 50.08  & 42.80\\ \hline
Human performance & - & 14.97 & - & - &-&  - & -\\ \hline

\end{tabular}
\label{tab:results}
}
\end{table*}

We note that segmentation results are not always directly comparable. For example, \newcite{chen2009global} require that all documents in the dataset discuss the same topic, and so their method is not directly applicable to \textsc{Wiki-50}. Nevertheless, we attempt a comparison in Table \ref{tab:results}.

% \jb{You did not define Choi to be a dataset}
%We compare our method to other methods in terms of segmentation quality and runtime. We report results on \textsc{Wiki-727k}, \textsc{Wiki-50}, Choi's synthetic benchmark, and the two small Wikipedia benchmarks from \cite{chen2009global}.

\subsection{Accuracy}

Comparing our method to \textsc{GraphSeg}, we can see that \textsc{GraphSeg} gives better results on the synthetic Choi dataset, but this success does not carry over to the natural Wikipedia data, where they underperform the random baseline. We explain this by noting that since the dataset is synthetic, and was created by concatenating unrelated documents, even the simple word counting method in \newcite{choi2000advances} can achieve reasonable success.
\textsc{GraphSeg} uses a similarity measure between word embedding vectors to surpass the word counting method, but in a natural document, word similarity may not be enough to detect a change of topic within a single document. At the word level, two documents concerning completely different topics are much easier to differentiate than two sections in one document.

We compare our method to \newcite{chen2009global} on the two small Wikipedia datasets from their paper. Our method outperforms theirs on \textsc{Cities} and obtains worse results on \textsc{Elements}, where presumably our word embeddings were of lower quality, having been trained on Google News, where one might expect that few technical words from the domain of Chemistry are used. We consider this result convincing, since we did not exploit the fact that all documents have similar structure as \newcite{chen2009global}, and did not train specifically for these datasets, but still were able to demonstrate competitive performance.

Interestingly, human performance on \textsc{Wiki-50} is only slightly better than our model. We assume that because annotators annotated only a small number of documents, they still lack familiarity with the right level of granularity for segmentation, 
and are thus at a disadvantage compared to the model that has seen many documents.

%and did not accumulate the prior knowledge about segmentation characteristics of specific topics. 

\subsection{Run Time}
Our method's runtime is linear in the number of words and the number of sentences in a document. Conversely, \textsc{GraphSeg} has a much worse asymptotic complexity of $O(N^3 + V^k)$ where $N$ is the length of the longest sentence, $V$ the number of sentences, and $k$ the largest clique size. Moreover, neural network models are highly parallelizable, and benefit from running on GPUs. 

In practice, our method is much faster than \textsc{GraphSeg}. 
In Table~\ref{run-time-table} we report the average run time per document on \textsc{Wiki-50} on a CPU.

\begin{table}[h]
\centering
{\footnotesize
\caption{Average run time in seconds per document.}
\label{run-time-table}
\begin{tabular}{|l|c|}
\hline & \textsc{Wiki-50}  \\ \hline
Our model (CPU) & 1.6     \\ \hline
\textsc{GraphSeg} (CPU) & 23.6  \\ \hline
\end{tabular}
}
\end{table}

\section{Conclusions}
In this work, we present a large labeled dataset, \textsc{Wiki-727k}, for text segmentation, that enables training neural models using supervised learning methods. This closes an existing gap in the literature, where thus far text segmentation models were trained in an unsupervised fashion.

Our text segmentation model outperforms prior methods on Wikipedia documents, and performs competitively on prior benchmarks. Moreover, our system has linear runtime in the text length, and can be run on modern GPU hardware. 
We argue that for text segmentation systems to be useful in the real world, they must be able to segment arbitrary natural text, and this work provides a path towards achieving that goal.

In future work, we will explore richer neural models at the sentence-level. Another important direction is developing a structured global model that will take all local predictions into account and then perform a global segmentation decision.

\section*{Acknowledgements} We thank  the anonymous reviewers
for their constructive feedback. This
work was supported by
the Israel Science
Foundation, grant 942/16.”
%With our new general, extensive, and natural dataset, we provide both a training set for such systems, and a benchmark for comparing them.
%While the model suggested in this work achieves good results, doubtlessly a more elaborate model will be able to achieve better results. We expect to see such results in the future.

%For a text segmentation system to be useful in the real world, it must be able to segment arbitrary natural text. With our new general, extensive, and natural dataset, we provide both a training set for such systems, and a benchmark for comparing them.

%Our baseline text segmentation model performs competitively with previous systems on restricted domain benchmarks. In addition, our system has linear runtime in the text size, and can be run very quickly on modern GPU hardware. While the model suggested in this work achieves good results, doubtlessly a more elaborate model will be able to achieve better results. We expect to see such results in the future.

\bibliography{cites}

\begin{thebibliography}{}
\expandafter\ifx\csname natexlab\endcsname\relax\def\natexlab#1{#1}\fi

\bibitem[{Beeferman et~al.(1999)Beeferman, Berger, and
  Lafferty}]{beeferman1999statistical}
Doug Beeferman, Adam Berger, and John Lafferty. 1999.
\newblock Statistical models for text segmentation.
\newblock {\em Machine learning\/} 34(1):177--210.

\bibitem[{Bird et~al.(2009)Bird, Loper, and Klein}]{bird2009nltk}
S.~Bird, E.~Loper, and E.~Klein. 2009.
\newblock {\em Natural Language Processing with Python\/}.
\newblock O’Reilly Media Inc.

\bibitem[{Blei et~al.(2003)Blei, Ng, and Jordan}]{blei03lda}
D.~Blei, A.~Ng, and M.~I. Jordan. 2003.
\newblock Latent {D}irichlet allocation.
\newblock {\em Journal of Machine Learning Research (JMLR)\/} 3:993--1022.

\bibitem[{Brants et~al.(2002)Brants, Chen, and
  Tsochantaridis}]{brants2002topic}
Thorsten Brants, Francine Chen, and Ioannis Tsochantaridis. 2002.
\newblock Topic-based document segmentation with probabilistic latent semantic
  analysis.
\newblock In {\em Proceedings of the eleventh international conference on
  Information and knowledge management\/}. ACM, pages 211--218.

\bibitem[{Chen et~al.(2009)Chen, Branavan, Barzilay, and
  Karger}]{chen2009global}
Harr Chen, SRK Branavan, Regina Barzilay, and David~R Karger. 2009.
\newblock Global models of document structure using latent permutations.
\newblock In {\em Proceedings of Human Language Technologies: The 2009 Annual
  Conference of the North American Chapter of the Association for Computational
  Linguistics\/}. Association for Computational Linguistics, pages 371--379.

\bibitem[{Choi(2000)}]{choi2000advances}
Freddy~YY Choi. 2000.
\newblock Advances in domain independent linear text segmentation.
\newblock In {\em Proceedings of the 1st North American chapter of the
  Association for Computational Linguistics conference\/}. Association for
  Computational Linguistics, pages 26--33.

\bibitem[{Fournier(2013)}]{Fournier2013a}
Chris Fournier. 2013.
\newblock {Evaluating Text Segmentation using Boundary Edit Distance}.
\newblock In {\em Proceedings of 51st Annual Meeting of the Association for
  Computational Linguistics\/}. Association for Computational Linguistics,
  Stroudsburg, PA, USA, page to appear.

\bibitem[{Glava{\v{s}} et~al.(2016)Glava{\v{s}}, Nanni, and
  Ponzetto}]{glavavs2016unsupervised}
Goran Glava{\v{s}}, Federico Nanni, and Simone~Paolo Ponzetto. 2016.
\newblock Unsupervised text segmentation using semantic relatedness graphs.
\newblock Association for Computational Linguistics.

\bibitem[{Hearst(1994)}]{hearst1994multi}
Marti~A Hearst. 1994.
\newblock Multi-paragraph segmentation of expository text.
\newblock In {\em Proceedings of the 32nd annual meeting on Association for
  Computational Linguistics\/}. Association for Computational Linguistics,
  pages 9--16.

\bibitem[{Hochreiter and Schmidhuber(1997)}]{hochreiter1997long}
Sepp Hochreiter and J{\"u}rgen Schmidhuber. 1997.
\newblock Long short-term memory.
\newblock {\em Neural computation\/} 9(8):1735--1780.

\bibitem[{Mikolov et~al.(2013)Mikolov, Yih, and Zweig}]{mikolov2013linguistic}
T.~Mikolov, W.~Yih, and G.~Zweig. 2013.
\newblock Linguistic regularities in continuous space word representations.
\newblock In {\em hlt-Naacl\/}. volume~13, pages 746--751.

\bibitem[{Pennington et~al.(2014)Pennington, Socher, and
  Manning}]{pennington2014glove}
Jeffrey Pennington, Richard Socher, and Christopher~D. Manning. 2014.
\newblock \href{http://www.aclweb.org/anthology/D14-1162}{Glove: Global vectors
  for word representation}.
\newblock In {\em Empirical Methods in Natural Language Processing (EMNLP)\/}.
  pages 1532--1543.
\newblock \url{http://www.aclweb.org/anthology/D14-1162}.

\bibitem[{Riedl and Biemann(2012)}]{riedl2012topictiling}
Martin Riedl and Chris Biemann. 2012.
\newblock Topictiling: a text segmentation algorithm based on lda.
\newblock In {\em Proceedings of ACL 2012 Student Research Workshop\/}.
  Association for Computational Linguistics, pages 37--42.

\bibitem[{Utiyama and Isahara(2001)}]{utiyama2001statistical}
Masao Utiyama and Hitoshi Isahara. 2001.
\newblock A statistical model for domain-independent text segmentation.
\newblock In {\em Proceedings of the 39th Annual Meeting on Association for
  Computational Linguistics\/}. Association for Computational Linguistics,
  pages 499--506.

\end{thebibliography}
\bibliographystyle{acl_natbib}

%\appendix
%\section{Appendix: Model Diagram}

\end{document}